\documentclass[twoside,11pt]{article}

\pdfoutput=1

\usepackage{graphicx}
\usepackage{amssymb}
\usepackage{caption}
\usepackage{subcaption}
\usepackage{mathtools}
\usepackage{amsmath}

\usepackage{jmlr2e}

\usepackage{hyperref}

\usepackage{float}
\usepackage[ruled]{algorithm2e}

\DeclareMathOperator*{\argmin}{\mathrm{arg\,min}}

\DeclareMathOperator*{\E}{\mathbb{E}}
\DeclareMathOperator*{\supp}{\mathrm{supp}}

\newcommand{\comment}[1]{}

\jmlrheading{x}{2019}{x-xx}{x/xx}{x/xx}{Borisyak19a}{Maxim Borisyak, Artem Ryzhikov, Andrey Ustyuzhanin, Denis Derkach, Fedor Ratnikov and Olga Mineeva}

\ShortHeadings{$(1 + \varepsilon)$-class classification}{Borisyak, Ryzhikov, Ustyuzhanin, Derkach, Ratnikov and Mineeva}
\firstpageno{1}

\binoppenalty=\maxdimen
\relpenalty=\maxdimen

\begin{document}

\title{$(1 + \varepsilon)$-class Classification: an Anomaly Detection Method for Highly Imbalanced or Incomplete Data Sets}

\author{\name Maxim Borisyak \email mborisyak@hse.ru\\
        \name Artem Ryzhikov \email aryzhikov@hse.ru\\
        \name Andrey Ustyuzhanin \email austyuzhanin@hse.ru\\
        \name Denis Derkach \email dderkach@hse.ru\\
        \name Fedor Ratnikov \email fratnikov@hse.ru\\
        \name Olga Mineeva \email omineeva@student.ethz.ch\\
        \addr Laboratory of Methods for Big Data Analysis\\
        National Research University Higher School of Economics\\
        20 Myasnitskaya ulitsa, Moscow 101000 Russia
}

\editor{Denis Derkach}

\maketitle

\begin{abstract}%   <- trailing '%' for backward compatibility of .sty file
Anomaly detection is not an easy problem since distribution of anomalous samples is unknown a priori. We explore a novel method that gives a trade-off possibility between one-class and two-class approaches, and leads to a better performance on anomaly detection problems with small or non-representative anomalous samples. The method is evaluated using several data sets and compared to a set of conventional one-class and two-class approaches.
\end{abstract}


%% Title, authors and addresses

\section{Introduction}
\label{sec:intro}

Monitoring of complex systems and processes often goes hand in hand with anomaly detection. Anomaly here means a representation of abnormal system behavior. Information on normal system behaviour is often available in abundance, compared to samples of abnormal behavior. In some cases the anomalies are rare, or distribution of anomalies is highly skewed. So given the high variability of anomalies, it leads to the fact that some types of anomalies are missing in the training data set. In other cases, when anomalous examples are obtained by means other than sampling target system, or when distribution of anomalies evolve over time, some types of anomalies might even be unknown in principle. A good realistic data set with anomalous behavior is provided by KDD-99 Cup \citep{KDD99}, with certain families of cyber-attacks present only in the test sample.

Conventional approaches for anomaly detection often involve one-class classification methods \citep{ocnn, svdd, doc, liu2008isolation, SVM},  which yield a soft boundary between the normal class region, and the rest of the feature space. Usually such methods are referred to as unsupervised, since those do not take into account labels of available data. As this piece of information might be important, those one-class methods potentially lead to the performance degradation for the cases with significant overlap between normal and abnormal samples in the feature space.

There is a rich profusion of two-class supervised classification methods that account for both class labels, leading to better results in the presence of labeled abnormal samples. However, those methods lack any guarantees for predictions outside of the regions of the feature space presented in the training data. It becomes especially problematic for incomplete anomalous samples, as a classifier might consistently make false-positive predictions for unseen anomalies.

\textbf{Contribution} \quad In this study, we develop a method that is aimed at combining the best of the two, one-class and two-class approaches, which we refer to as $(1 + \varepsilon)$-class classification (\textit{'one plus epsilon'} or \textit{OPE} for short).
In order to achieve that, we derive two one-class objectives and combine them with the binary cross-entropy loss.
We compare these objectives with respect to computational effectiveness, and demonstrate performance on several data sets that are either collected for anomaly detection tasks \citep{KDD99}, or artificially under-sampled to emulate these conditions \citep{higgs, mnist, cifar, omniglot}.

\textbf{Notation} \quad We assume that an N-dimensional feature space $\mathcal{X}$ ($\mathbb{R}^{N}$), contains samples of two classes: normal (positive) $\mathcal{C}^+$ and abnormal (negative) $\mathcal{C}^-$. We are interested in identifying instances of the single class $\mathcal{C}^+$. There are two principal approaches: one-class (unitary classification) and two-class (binary classification). The former might rely on estimation of the likelihood of the positive class $P(x \mid \mathcal{C^+})$, so then one can apply a threshold to make a final decision. We will refer to any solution of the form $s(P(x \mid \mathcal{C^+}))$, where $s: \mathbb{R} \to \mathbb{R}$ is a monotonous function, as a \textit{unitary classification solution}. The latter relies on estimation of the posterior conditional distribution $P(\mathcal{C^+} \mid x)$, that is usually approximated through minimization of the cross-entropy loss function:
\begin{equation}
    \mathcal{L}_2(f) =
    P(\mathcal{C}^+) \E_{x \sim \mathcal{C}^+} \log f(x) + P(\mathcal{C}^-) \E_{x \sim \mathcal{C}^-} \log \left(1 - f(x) \right);
    \label{eq:loss1}
\end{equation}

where $\E_{x \sim \mathcal{C}} h(x)$ denotes conditional expectation $\E_{x}\left[ h(x) \mid \mathcal{C} \right]$, and $f : \mathcal{X} \to [0, 1]$ --- classifier's decision function.

Optimal binary decision function $f^*$ that minimizes $\mathcal{L}_2(f)$, can be expressed with the help of Bayes' rule as
\begin{equation}
    f^*(x) = P(\mathcal{C}^+ \mid x) = \frac{P(x \mid \mathcal{C}^+) P(\mathcal{C}^+)}{P(x \mid \mathcal{C}^+)P(\mathcal{C}^+) + P(x \mid \mathcal{C}^-)P(\mathcal{C}^-)}; \label{eq:l0solution}
\end{equation}
where $P(\mathcal{C})$ is class prior probability, $P(\mathcal{C} \mid x)$ --- posterior conditional distribution and $P(x \mid \mathcal{C})$ --- likelihood for the given class $\mathcal{C}$.

\section{One plus epsilon method}
\label{sec:method}
Let's consider a simple case: $\mathcal{C}^-$ is a uniform distribution $U[\Omega]$, with the support $\Omega$ covering that of $P(x \mid \mathcal{C}^+)$.
If we put this $\mathcal{C}^-$ into Equation~\ref{eq:loss1} (assuming equal class priors), we get
\begin{eqnarray}
    \mathcal{L}_1(f) &=& -\frac{1}{2} \left[ \E_{x \sim \mathcal{C}^+} \log f(x) + \E_{x \sim U[\Omega]} \log (1 - f(x)) \right];\nonumber \\
    f^*_1(x) &=& \argmin_f \mathcal{L}_1(f) =  \frac{P(x \mid \mathcal{C}^+)}{P(x \mid \mathcal{C}^+) + C}; \nonumber
\end{eqnarray}
where $C$ --- probability density of distribution $U[\Omega]$.
Note, that $f^*_1(x)$ is a unitary classification solution, therefore, \textit{solution to a classification problem between a given class and a uniformly distributed one, yields a unitary classification solution}.

\subsection{Adding known negative samples}
Let's take into account known anomalous samples. We propose the following loss function --- linear combination of one-class classification loss $\mathcal{L}_1$ and cross-entropy loss $\mathcal{L}_2$:
\begin{eqnarray}
    \mathcal{L}_{1 + \varepsilon}(f) &=&
    \frac{1}{2} \left( L^+(f) + \gamma\,L^-(f) + (1 - \varepsilon)\,L^0(f) \right); \label{eq:opeloss}\\
    L^+(f) &=& -\E_{x \sim \mathcal{C}^+} \log f(x); \nonumber\\
    L^-(f) &=& -\E_{x \sim \mathcal{C}^-} \log (1 - f(x)); \nonumber\\
    L^0(f) &=& -\E_{x \sim U} \log (1 - f(x)); \nonumber
\end{eqnarray}
where $\gamma$ compensates for the difference in classes prior probabilities. Ideally, it should be set to $P(\mathcal{C}^-) / P(\mathcal{C}^+)$, so that the first two terms match the cross-entropy loss. $\varepsilon$ is a hyper-parameter, that allows to choose the trade-off between unitary and binary classification solutions. We call the loss $\mathcal{L}_{1 + \varepsilon}(f)$  \textit{OPE loss}.
It leads to the following solution:
\begin{equation}
    f^*_{1 + \varepsilon}(x) = \frac{P(x \mid \mathcal{C}^+)}{P(x \mid \mathcal{C}^+) + (1 - \varepsilon)\,C + \gamma\,P(x \mid \mathcal{C}^-)} \nonumber.
\end{equation}

\begin{figure}[tbp]
    \centering
    \begin{subfigure}[b]{0.3\textwidth}
        \includegraphics[width=\textwidth]{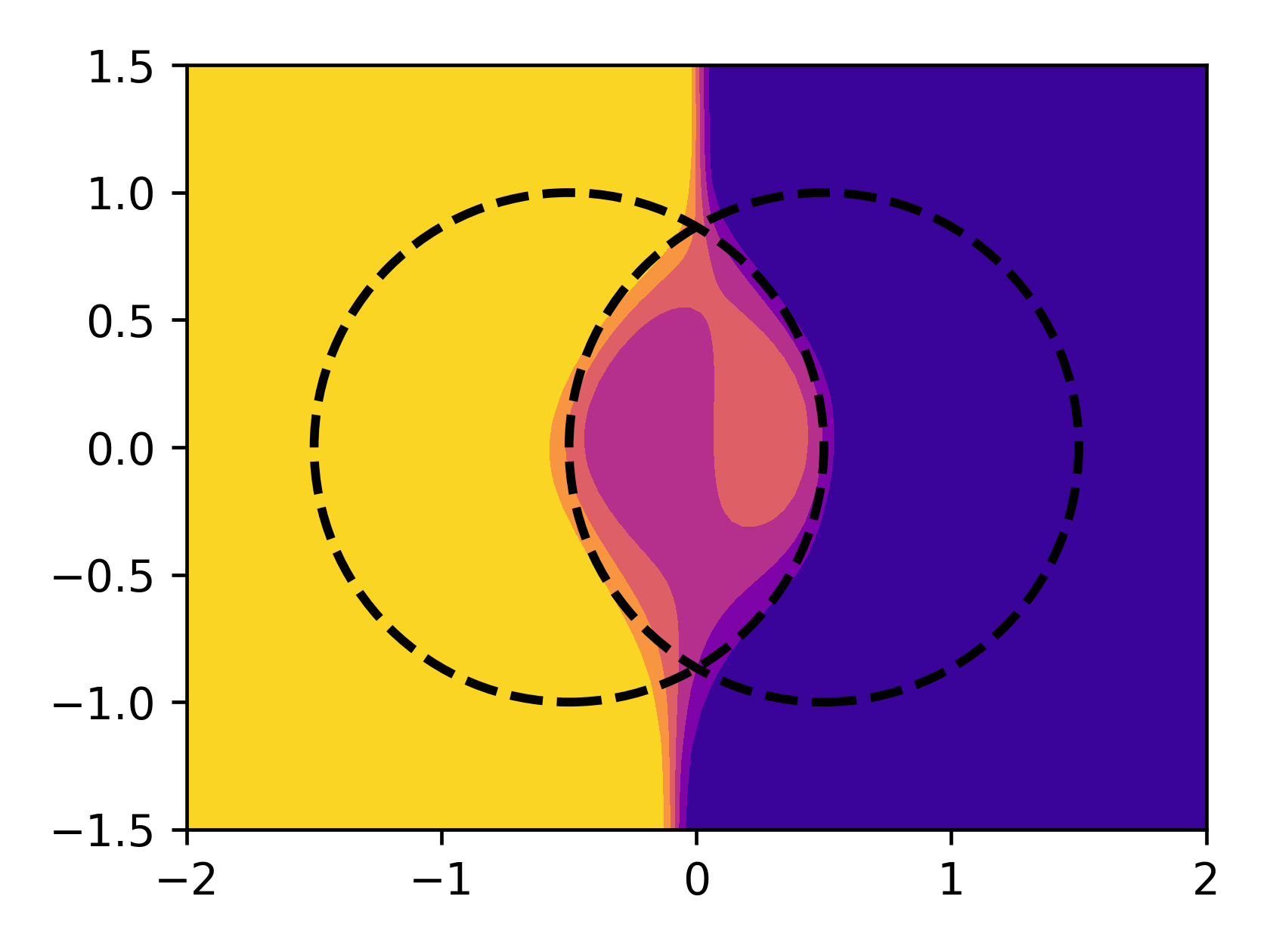}
        \caption{two-class classification}
    \end{subfigure}
    ~
    \begin{subfigure}[b]{0.3\textwidth}
        \includegraphics[width=\textwidth]{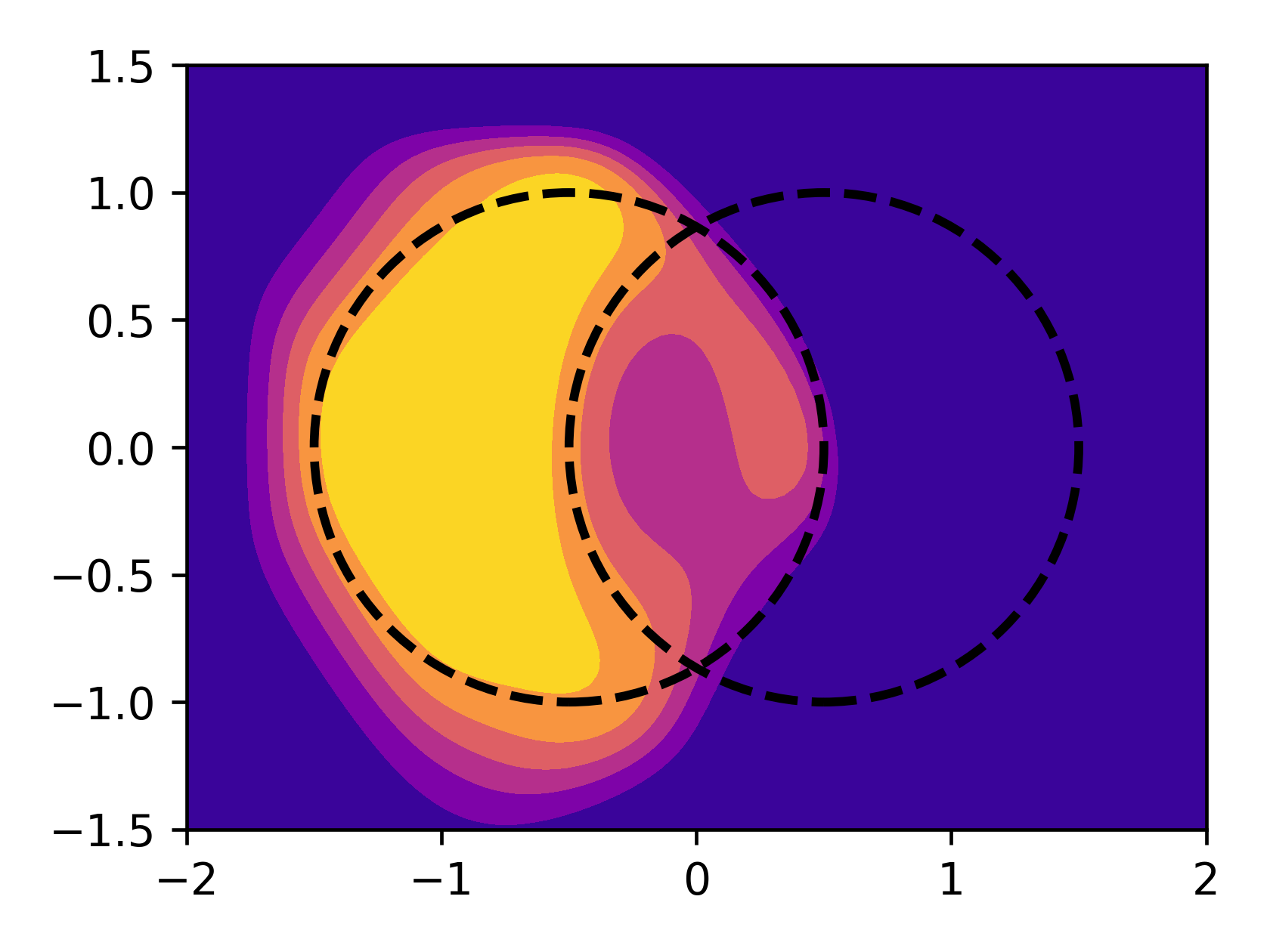}
        \caption{OPE classification}
    \end{subfigure}
    ~
    \begin{subfigure}[b]{0.3\textwidth}
        \includegraphics[width=\textwidth]{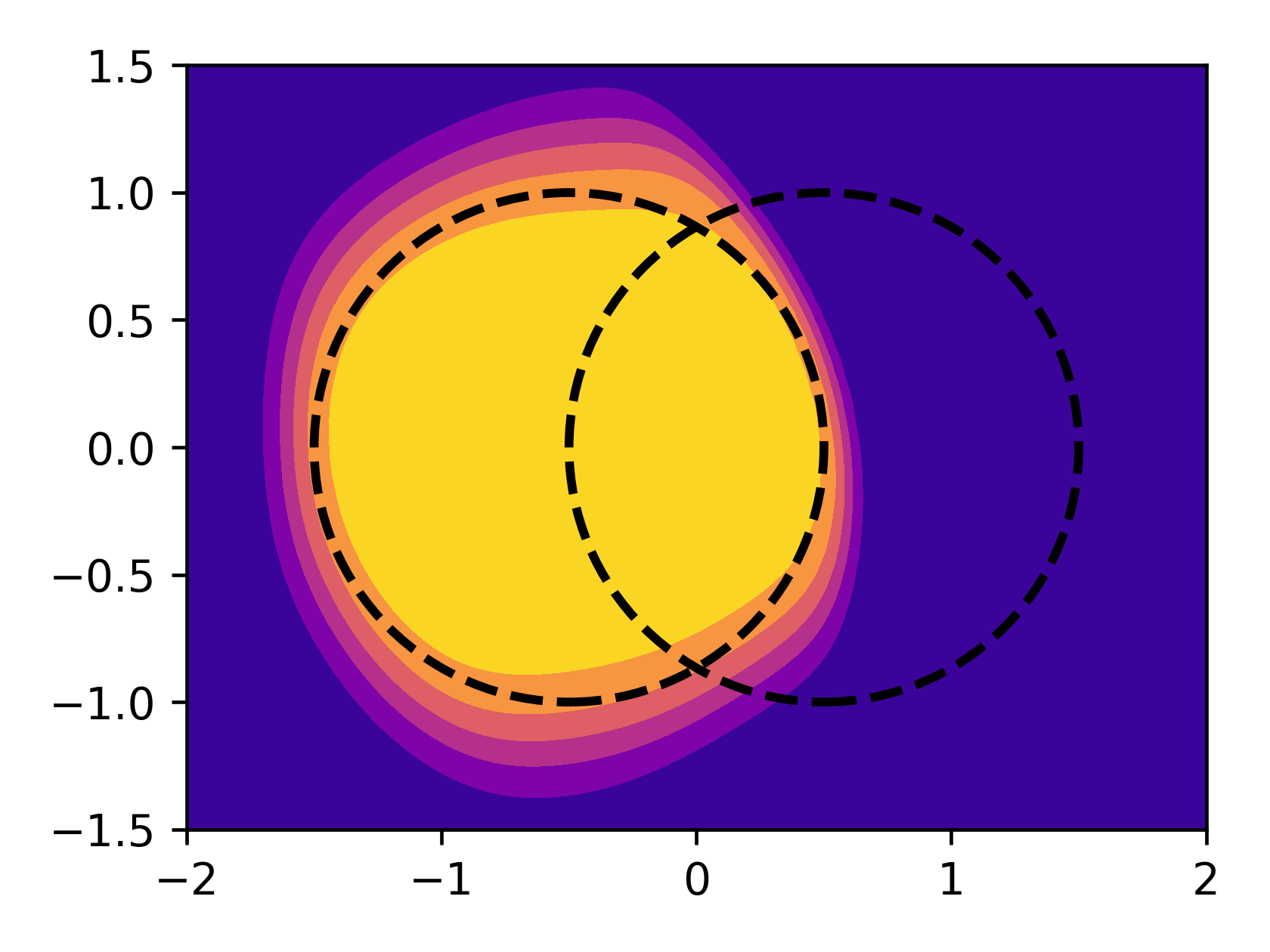}
        \caption{one-class classification.}
    \end{subfigure}
    \caption{Demonstration of the main idea behind OPE loss. Samples are uniformly distributed within areas bounded by the circles: the left one as positive class, the right one as negative. One-class solution was obtained by setting $\gamma = 1$, and $\varepsilon = 0$. Training samples are not shown for visual clarity.}
    \label{fig:opedemo}
\end{figure}

An important observation can be made --- for large capacity models, even $\varepsilon$ close to $1$, leads to a significantly different solution in comparison to  the two-class classification solution (Equation~\ref{eq:l0solution}). This effect can be seen on Figure \ref{fig:opedemo}.

One might consider the term $L_0$ as a regularization term, that biases solution $f^*_{1 + \varepsilon}(x)$ towards 0 everywhere, but this effect is especially pronounced in points with $P(x \mid \mathcal{C}^+) \approx 0$. One distinguishing feature of the $L_0$ regularization term is that it acts directly on predictions, rather then on parameters\footnote{Technically, regularization in one-class SVM objective \citep{scholkopf2000support} and similar methods can also be considered to act directly on predictions since these are linear models.}, which makes it applicable to any classifier model.

Estimating $L^0$ term in Equation~\ref{eq:opeloss} for low-dimensional feature-space is straightforward --- if $\supp P(x \mid \mathcal{C}^+)$ can be bounded by a simple set $\Omega$ (e.g. a box), then $L_0$ can be estimated by directly sampling from $U[\Omega]$. We refer to this class of OPE algorithms as \textit{brute-force OPE}.

\subsection{Energy-based regularization}
\label{sec:eope}
For high-dimensional feature space, however, sampling directly from $U[\Omega]$ might be problematic, due to a potentially high variance of the gradients produced by the regularization term. One possible strategy of reducing variance of $L^0$ gradient estimates, is to sample from another distribution $Q$:
\begin{equation}
    L^0(f) = \E_{x \sim U} \log(1 - f(x)) = \E_{x \sim Q} \frac{C}{Q(x)} \log(1 - f(x)) \label{eq:isope};
\end{equation}

\cite{icn} employs this method and uses distribution $Q = P_f$ induced by the model $f$ at the previous training epoch:
\begin{equation}
    P_f(x) = \frac{1}{Z} \frac{\mathbb{I}[x \in \Omega] \cdot f(x)}{1 - f(x)}; \nonumber
\end{equation}
where: $Z = \int_{\Omega} \frac{f(x)}{1 - f(x)} dx$ --- normalization term, $\mathbb{I}$ --- indicator function. Hence, $L^0$ can be written as:
\begin{equation}
    L^0(f) = Z \cdot \E_{x \sim P_f} C \cdot \frac{1 - f(x)}{f(x)} \log(1 - f(x)) \nonumber.
\end{equation}

Sampling from $P_f$ is computationally expensive, and various methods can be used, e.g. Hamiltonian Monte-Carlo \citep{hmc}.
However, this transformation merely transfers the computationally heavy integration part from uniform sampling to estimation of the normalization term $Z$. In order to avoid recomputing $Z$ on each epoch, a two-stage training procedure is proposed by \citet{icn} and \citet{dgl}:
\begin{enumerate}
    \item freeze sampling distribution $P_f$, estimate $Z$;
    \item using this frozen distribution perform a number of stochastic gradient descent steps.
\end{enumerate}

Note, that as long as a regularization term shifts the decision function towards $0$ outside of $\supp P(x \mid \mathcal{C}^+)$, and has a small impact within, it suffices for the purposes of anomaly detection. With that idea in mind, we propose the following approximation of $L^0$ regularization term to avoid uniform sampling and integration.

Let's introduce $g(x) = \sigma^{-1}(f)$, where $\sigma(\chi) = 1 / \left[ 1 + \exp(-\chi) \right]$ --- sigmoid function:
\begin{eqnarray}
    P_g(x) &=& \frac{1}{Z}\frac{f(x)}{1 - f(x)} = \frac{1}{Z} \exp(g(x)); \label{eq:energydef}\\
    \text{where} \medspace Z &=& \int_{\Omega} \exp(g(x)) dx \nonumber.
\end{eqnarray}

Note, that $g(x)$ in Equation~\ref{eq:energydef} matches the definition of (negative) energy $E(x)$ used in energy-based generative models \citep{bengio2009learning}: $P(x) \propto \exp(-E(x))$.

In case of $Z \gg C$, using Jensen inequality, we can approximate upper bound of $L^0$ as follows:
\begin{multline}
    L^0 =  -\E_{x \sim U} \log(1 - f(x)) = \\
    \E_{x \sim U} \log(1 + \exp(g(x))) \leq \log \left[ 1 + \E_{x \sim U} \exp(g(x)) \right] = \\
    \log\left(1 + \frac{Z}{C}\right) \approx \log Z - \log C; \nonumber
\end{multline}
which leads to the following one-class loss function:
\begin{eqnarray}
    \mathcal{L}^E_1(g) &=& \frac{1}{2}\left[ \E_{x \sim \mathcal{C}^+} \log\left(1 + \exp(-g(x))\right) + (1 - \varepsilon) L^E(g) \right]; \label{eq:eocloss}\\
    \text{where} \medspace L^E(g) &=& \log Z = \int_{\Omega} \exp(g(x)) dx; \nonumber
\end{eqnarray}
then the corresponding energy OPE (EOPE) loss function is
\begin{eqnarray}
    \mathcal{L}^E_{1 + \varepsilon}(f) = \frac{1}{2}\left(L^+(f) + \gamma L^-(f) + (1 - \varepsilon) L^E(\sigma^{-1}(f))\right).
    \label{eq:eopeloss}
\end{eqnarray}

Gradients of $L^0_E$ can be easily estimated (see e.g. \citet{bengio2009learning}):
\begin{equation}
    \nabla L^E(g) = \nabla \log Z = \frac{1}{Z} \int_{\Omega} \exp(g(x)) \nabla g(x) = \E_{x \sim P_g} \nabla g(x) \label{eq:energygrad}.
\end{equation}

Note, that Equation~\ref{eq:energygrad} essentially describes the negative phase of contrastive divergence algorithm for energy-based models. Similar relations between the cross-entropy loss and contrastive divergence have also been mentioned by \citet{kim2016deep}.

As discussed above, the main goal of $L^0$ regularization term is to enforce one-class properties, namely, make the solution to be a monotonous transformation of $P(x \mid  \mathcal{C}^+)$. The following theorem shows that, despite being just an approximation of $L^0$ regularization, \textit{$\mathcal{L}^E_1(g)$ loss always leads to a one-class solution}.

{\bf Theorem 1} {\it
    Let $(\mathcal{X}, \|\cdot\|)$ be a Banach space, $P(x)$ --- a continuous probability density function such that $\Omega = \supp P$ is an open set in $\mathcal{X}$.
    If continuous function $g^* : \Omega \to \mathbb{R}$ minimizes $\mathcal{L}^E_1$ (defined by Equation~\ref{eq:eocloss}) with $P(x \mid \mathcal{C}^+) = P(x)$, then there exists a strictly increasing function $s: \mathbb{R} \to \mathbb{R}$, such that $g^*(x) = s(P(x))$. Moreover, $\lim_{y \to 0} s(y) = -\infty$ (if $\inf_\Omega P = 0$).
} \hfill\BlackBox

\noindent
Intuitively, it is clear, that if the dependency between $g(x)$ and $P(x)$ is violated in some regions, energy can be exchanged between these regions with a total reduction in the loss. A similar argument can be made for the property: $\lim_{y \to 0} s(y) > -\infty$ --- energy of low-density regions can be transferred to a high-density region, leading to an improved solution. A more formal proof can be found in Appendix B.

\section{Implementation details}

While OPE and EOPE losses are independent from any particular choice of model $f$, we consider only neural networks. We optimize all neural networks with a stochastic gradient method (namely, adam algorithm by to \citet{kingma2014adam}).
Algorithms \ref{algo:ope} and \ref{algo:eope} outline proposed methods.

\begin{algorithm}[tbp]
    \SetKwInput{KwHyper}{Hyper-parameters}
    \caption{Brute-force OPE}
    \label{algo:ope}
    \KwIn{
    $\mathrm{normal\;data}$, $\mathrm{anomalous\;data}$ --- samples from $\mathcal{C}^+$, $\mathcal{C}^-$, the latter might be absent; $f_\theta$ --- a classifier with parameters $\theta$.
    }
    \KwHyper{
        $\gamma$ --- ratio of class priors;
        $\varepsilon$ --- controls strength of regularization.
    }
    \While{not converged}{
        sample normal data $\{ x^+_i \sim \mathrm{normal\;data} \}^m_{i = 1}$\;
        sample known anomalies $\{ x^-_i \sim \mathrm{anomalous\;data} \}^m_{i = 1}$\;
        sample negative examples $\{ x^0_i \sim U[\Omega] \}^m_{i = 1}$\;
        $\nabla L^+ \leftarrow -\sum_i \nabla_\theta \log f_\theta(x^+_i)$\;
        $\nabla L^- \leftarrow -\sum_i \nabla_\theta \log(1 - f_\theta(x^-_i))$\;
        $\nabla L^0 \leftarrow -\sum_i \nabla_\theta \log(1 - f_\theta(x^0_i))$\;
        $\theta \leftarrow \mathrm{adam}\left(\nabla L^+ + \gamma \nabla L^- + (1 - \varepsilon)\nabla L^0\right)$
    }
\end{algorithm}

Estimation of $L^E(f)$ is tightly linked to the negative phase of energy-based generative models.
A traditional approach for sampling from $P_f$ is to employ Monte-Carlo (MC) methods, in this work we use Hamiltonian Monte-Carlo (HMC). Additionally, in our experiments we use persistent MC chains following \citet{tieleman2008training}. Nevertheless, usage of MC leads to a significant slow down of the training procedure, as in general, multiple passes through the network are required for generating negative samples.

Note, that for values of $\varepsilon$ close to 1, both $L^0$ and $L^E$ have a significant impact only in the regions with low probability density $P(x \mid \mathcal{C}^+)$. This suggests that solutions of Equations~\ref{eq:opeloss}~and~\ref{eq:eopeloss} are relatively robust to improper sampling procedures, and one might achieve a faster training without sacrificing much of quality, by employing fast approximate MC procedures. In our experiments we observed that the following highly degenerate instance of HMC is performing well:
\begin{eqnarray}
    x_{t + 1} &=& x_t + \eta \left[ \frac{\nabla g(x)}{\sqrt{m_{t + 1}}} + \lambda \xi_t \right];\label{eq:rmsprop1}\\
    m_{t + 1} &=& \rho m_t + (1 - \rho) (\nabla g(x) \odot \nabla g(x));\label{eq:rmsprop2}
\end{eqnarray}
where: $\odot$ denotes Hadamard product, $\xi_t$ is distributed normally with zero mean and unit covariance matrix, $\lambda > 0$ controls the impact of the random noise. We refer to the methods utilising such sampling as RMSProp-EOPE, since the procedure resembles RMSProp optimization algorithm \citep{rmsprop}. 

\begin{algorithm}[tp]
    \SetKwInput{KwHyper}{Hyper-parameters}
    \caption{Energy OPE}
    \label{algo:eope}
    \KwIn{
    $\mathrm{normal\;data}$, $\mathrm{anomalous\;data}$ --- samples from $\mathcal{C}^+$, $\mathcal{C}^-$, the latter might be absent; $g_\theta$ --- a classifier with parameters $\theta$.
    }
    \KwHyper{
        $\gamma$ --- ratio of class priors;
        $\varepsilon$ --- controls strength of regularization;
        $\mathrm{MCMC}$ --- Monte-Carlo sampling procedure.
    }
    \While{not converged}{
        sample normal data $\{ x^+_i \sim \mathrm{normal\;data} \}^m_{i = 1}$\;
        sample known anomalies $\{ x^-_i \sim \mathrm{anomalous\;data} \}^m_{i = 1}$\;
        sample negative examples $\{ x^0_i \sim \mathrm{MCMC}\left[x \mapsto \exp(g(x))\right] \}^m_{i = 1}$\;
        $\nabla L^+ \leftarrow \sum_i \nabla_\theta \log (1 + \exp(-g_\theta(x^+_i))$\;
        $\nabla L^- \leftarrow \sum_i \nabla_\theta \log(1 + \exp(g_\theta(x^-_i)))$\;
        $\nabla L^E \leftarrow \sum_i \nabla_\theta g_\theta(x^0_i)$\;
        $\theta \leftarrow \mathrm{adam}\left(\nabla L^+ + \gamma \nabla L^- + (1 - \varepsilon)\nabla L^E\right)$
    }
\end{algorithm}

A completely different approach to negative phase sampling is described by \citet{kim2016deep}. The authors suggest using a separate network (generator) to produce samples from the target distribution. We also implement this sampling procedure and refer to the methods employing it as Deep EOPE.

In our experiments, we observe that methods based on EOPE loss, quickly lead to steep functions which heavily interferes with the sampling procedures. Following \citet{prbm}, we add a small $l_2$ regularization term for predictions in pseudo-negative points:
\begin{equation}
    \tilde{L}^{E}(g) = \int_{\Omega} \exp(g(x)) dx + c \, \E_{x \sim P_f} \|g(x)\|^2; \nonumber
\end{equation}
where $c$ is a small constant ($c = 10^{-3}$ in our experiments).

\begin{figure}[tbp]
    \centering
    \begin{subfigure}[b]{0.48\textwidth}
        \includegraphics[width=\textwidth]{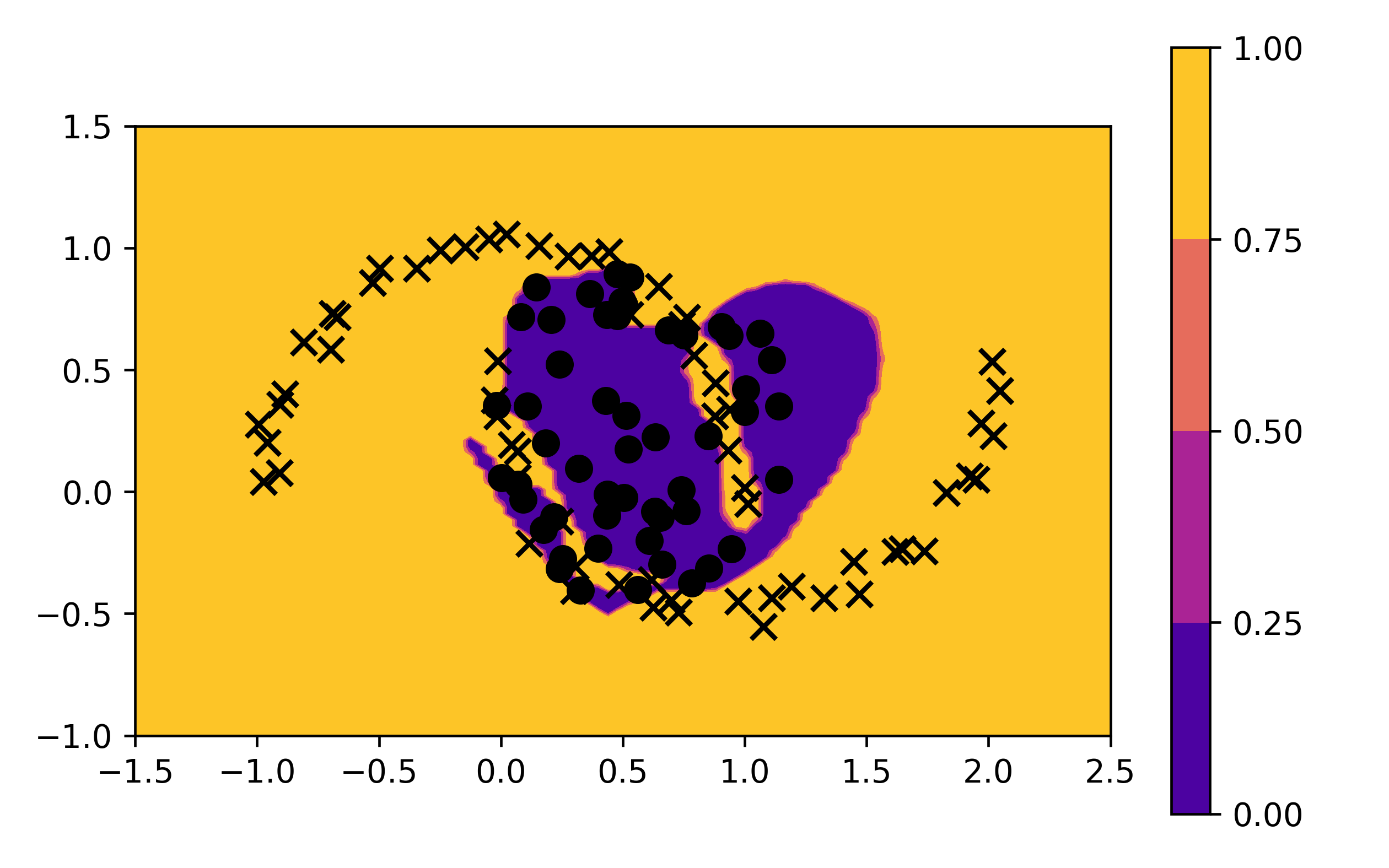}
        \caption{Two-class classification}
        \label{fig:twoclass}
    \end{subfigure}
    ~
    \begin{subfigure}[b]{0.48\textwidth}
        \includegraphics[width=\textwidth]{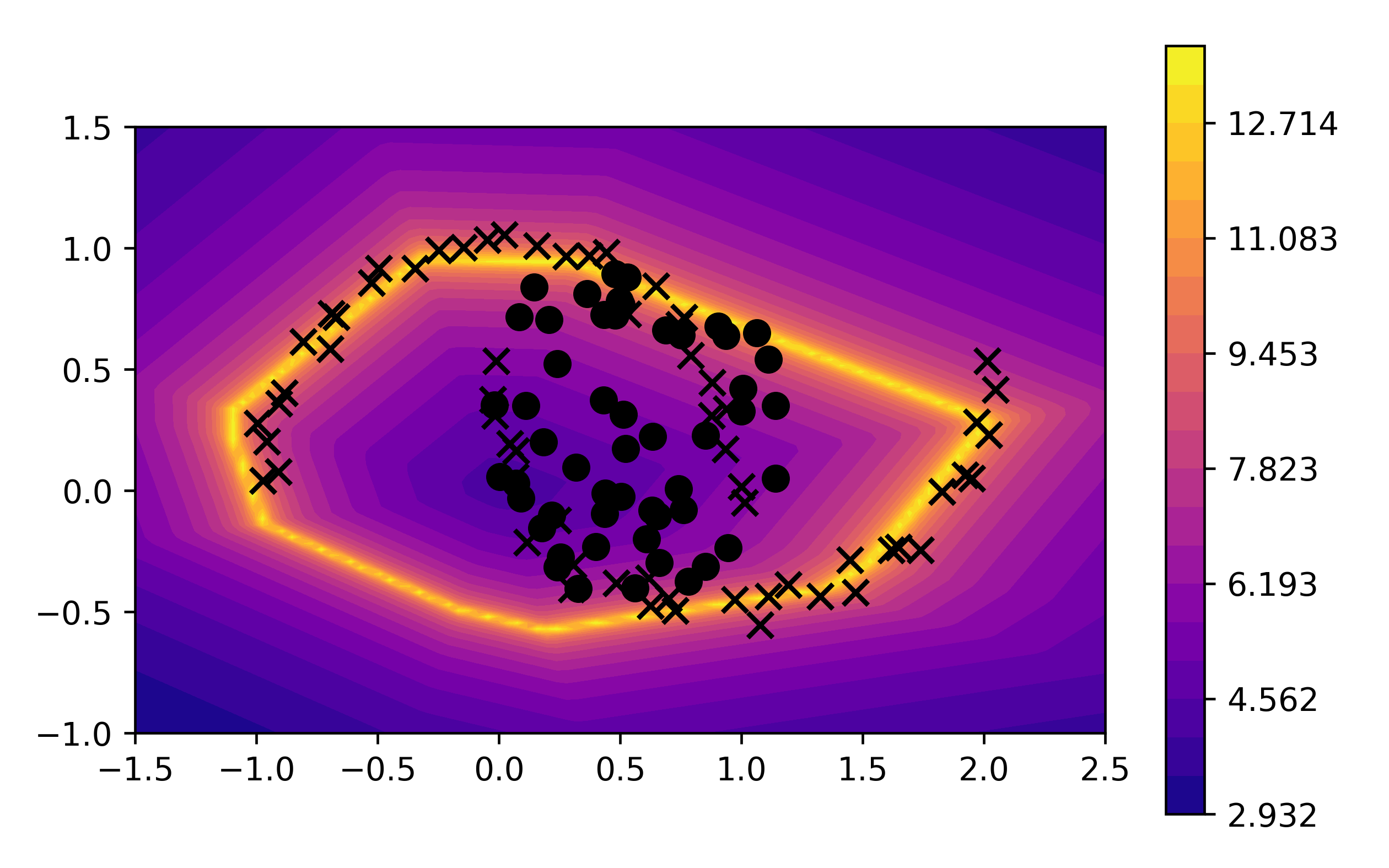}
        \caption{Deep SVDD}
        \label{fig:svdd}
    \end{subfigure}
    \\
    \begin{subfigure}[b]{0.48\textwidth}
        \includegraphics[width=\textwidth]{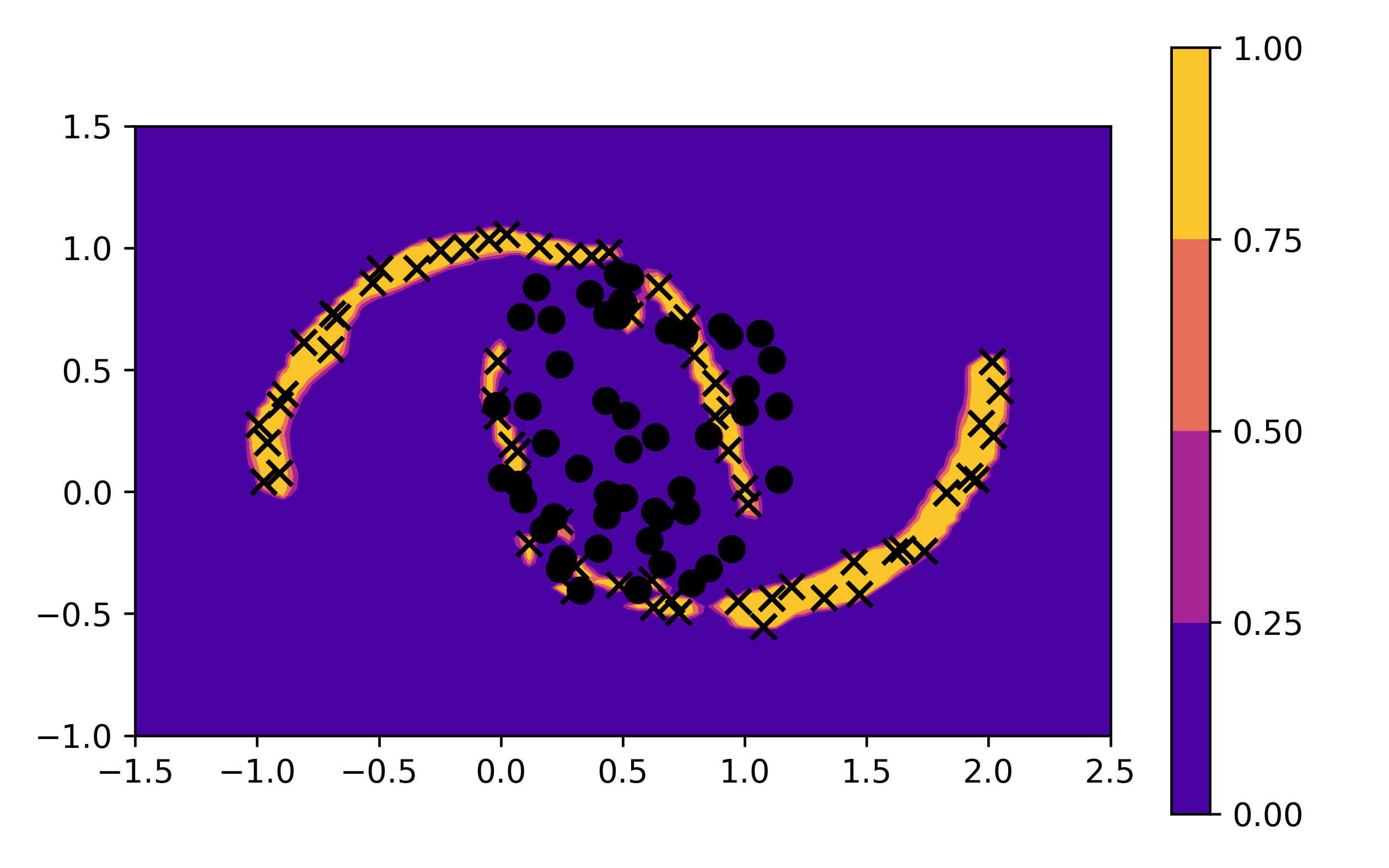}
        \caption{Brute-force OPE}
        \label{fig:bfope}
    \end{subfigure}
    ~
    \begin{subfigure}[b]{0.48\textwidth}
        \includegraphics[width=\textwidth]{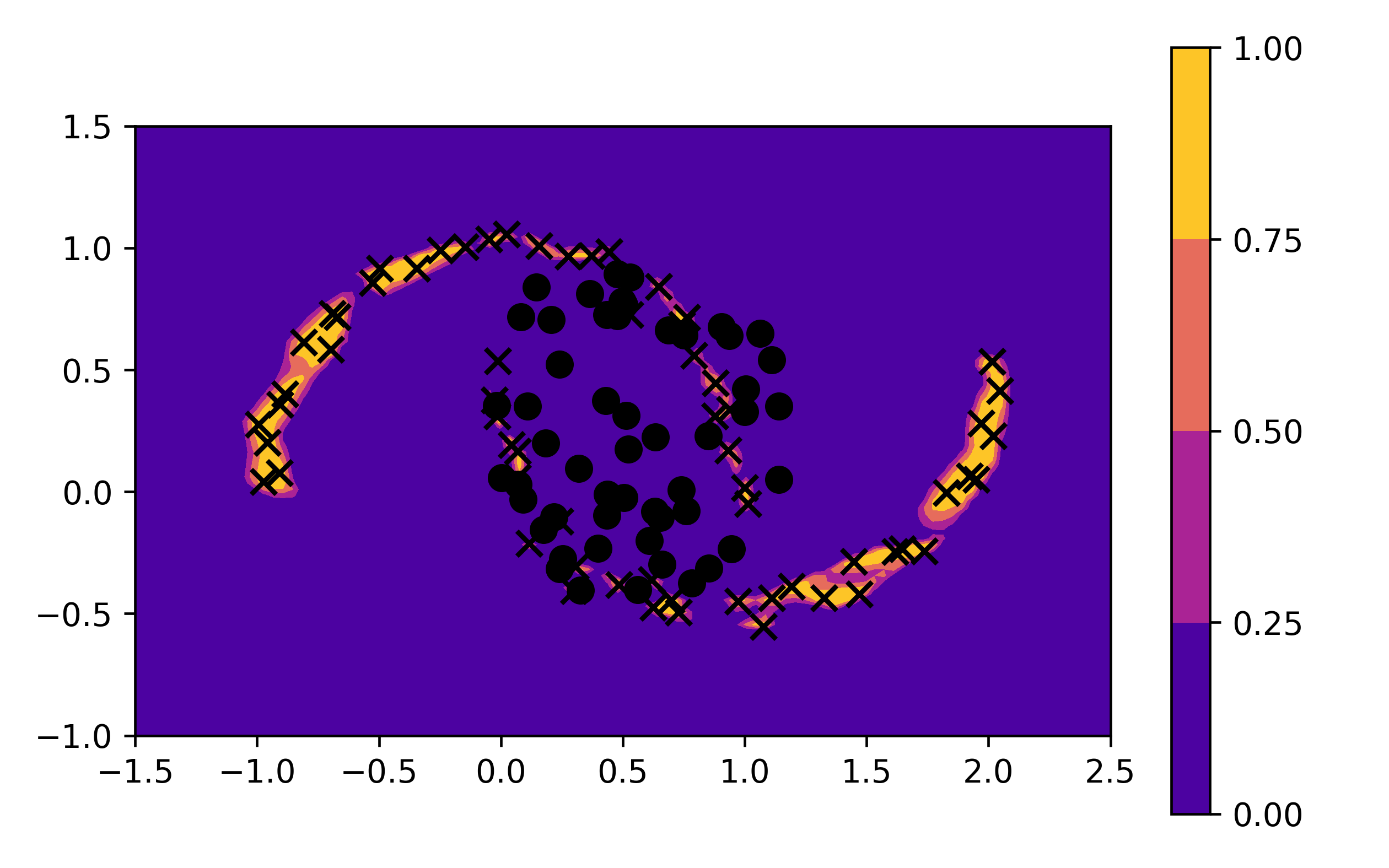}
        \caption{HMC EOPE}
        \label{fig:hmceope}
    \end{subfigure}
    \\
    \begin{subfigure}[b]{0.48\textwidth}
        \includegraphics[width=\textwidth]{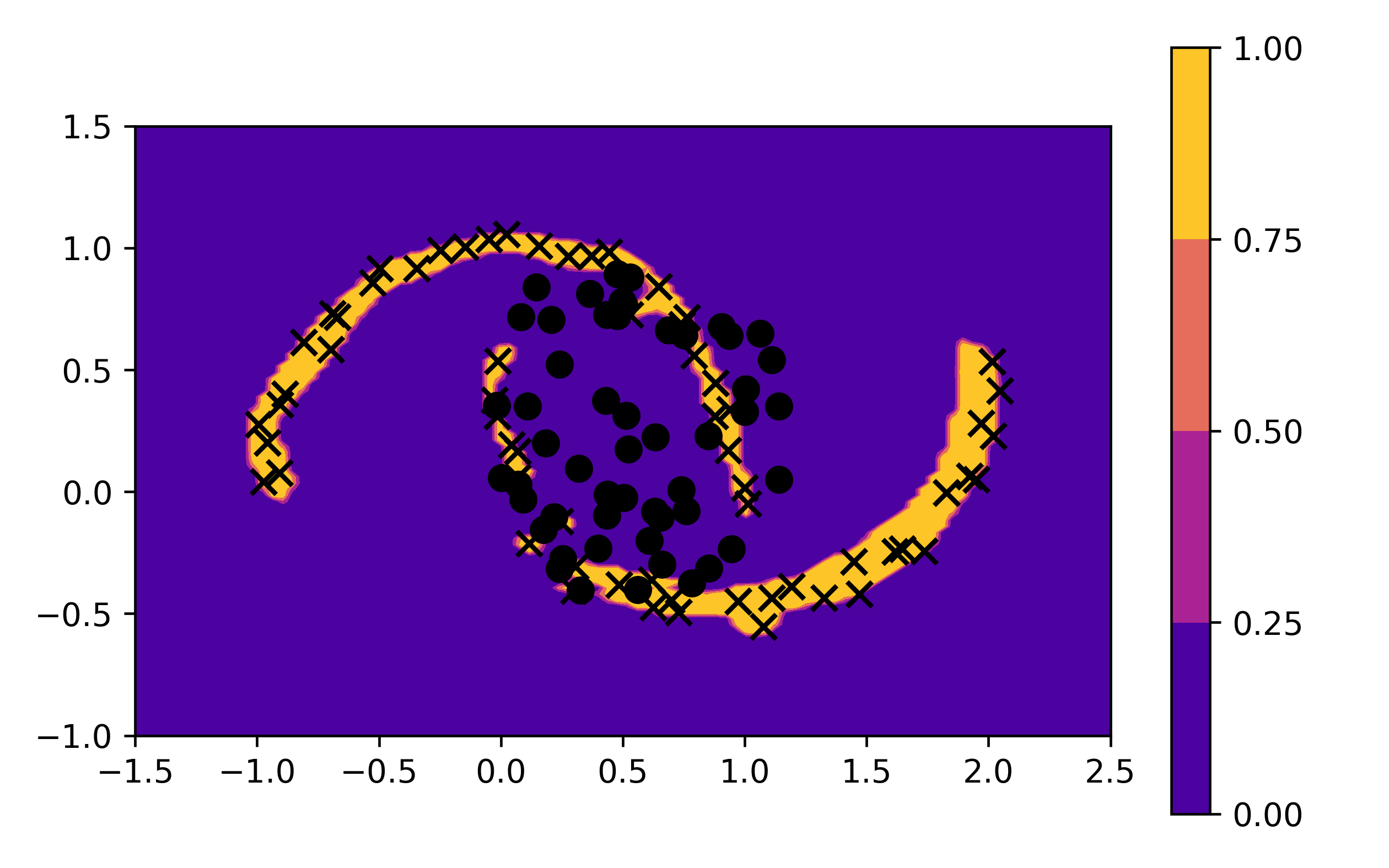}
        \caption{RMSProp EOPE}
        \label{fig:rmspropeope}
    \end{subfigure}
    ~
    \begin{subfigure}[b]{0.48\textwidth}
        \includegraphics[width=\textwidth]{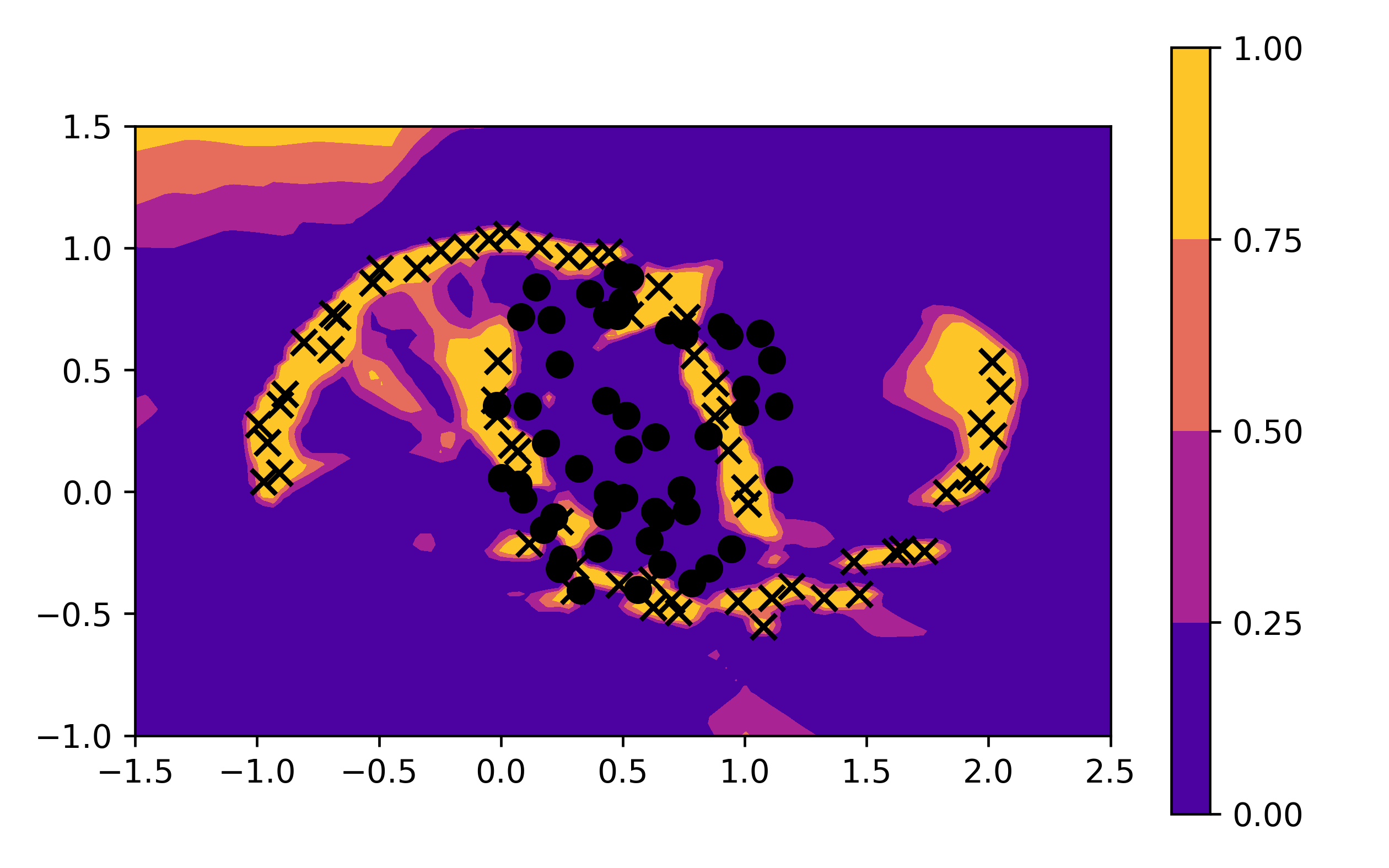}
        \caption{Deep EOPE}
        \label{fig:deepeope}
    \end{subfigure}
    
    \caption{Comparison of different methods on a toy example: positive examples (marked as 'x') are sampled from the Moons data set, negative examples (marked by black circles) are sampled uniformly from a circle of radius $\frac{1}{2}$. For visual consistency negative logarithm of Deep SVDD output is displayed.}
    \label{fig:moons}
\end{figure}

\begin{figure}[tbp]
    \includegraphics[width=\textwidth]{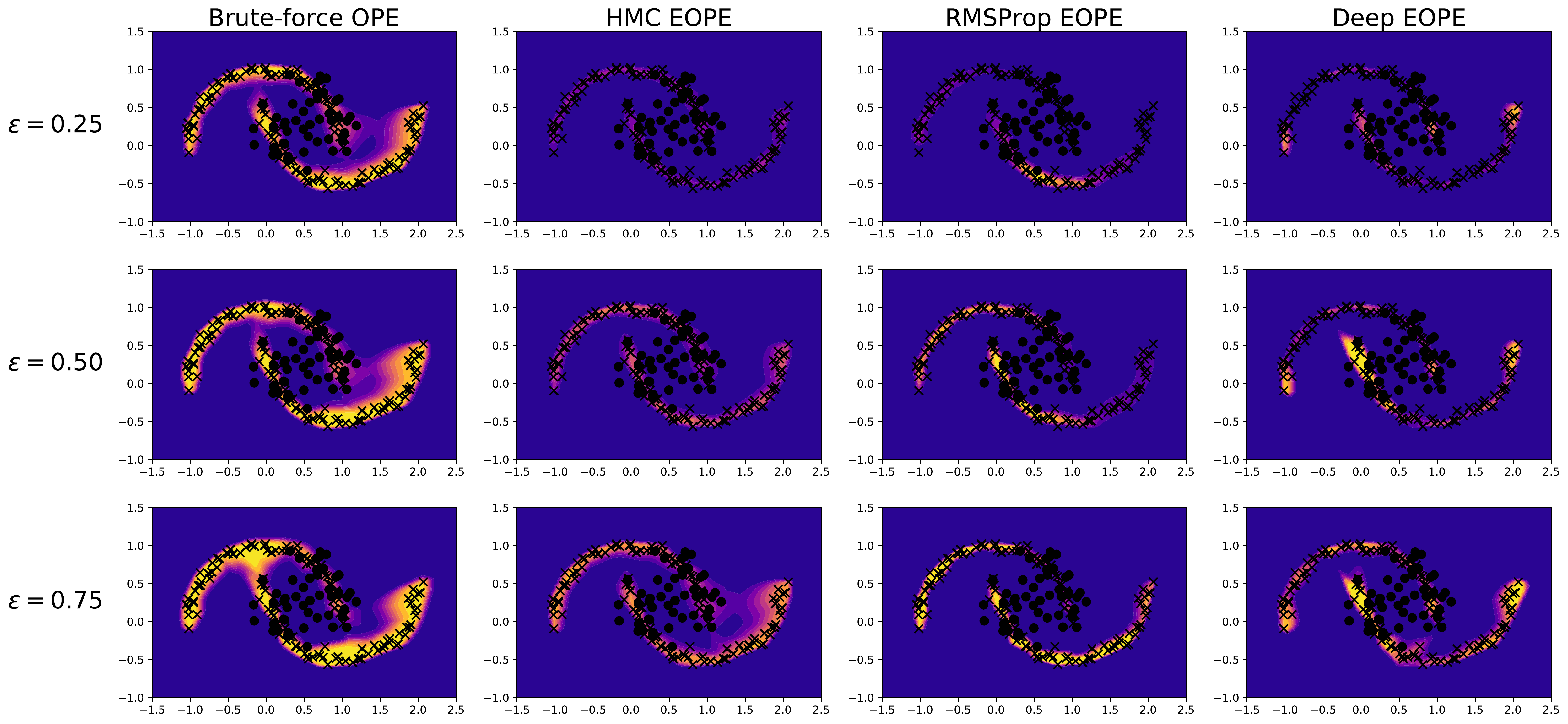}
    \caption{Comparison of OPE and EOPE losses with varying $\varepsilon$. For $\varepsilon < 1$, all losses lead to similar solutions. It appears that EOPE loss tends to overpenalize large predictions in contrast to OPE loss.}
    \label{fig:moonsepsvar}
\end{figure}

Figures \ref{fig:moons} and \ref{fig:moonsepsvar} demonstrate results of proposed methods on a toy data set.

\section{Relation to other methods}

The idea to perform one-class classification (and generative task) as 'one against everything', appears in many studies. \cite{svdd} propose constructing a hyper-sphere around positive samples, effectively separating it from the rest of the space; \citet{doc} and \citet{ocnn} extend this idea on deep neural networks. \cite{doc} rely on weight regularization, which acts in a similar manner to EOPE by limiting the area with high model output. OPE and EOPE methods depend only on the model's output, which allows OPE and EOPE methods to avoid limiting number of layers \citep[for example,][]{ocnn}, and does not restrict choice of network architecture \citep{doc}.

\citet{dgl} and \citet{icn} developed a method similar in its nature to OPE, in fact, it is easy to see, that $L_0$ term as it appears in Equation~\ref{eq:isope}, corresponds to the loss function from \cite{icn}. In this work we demonstrate that this loss is equivalent to the cross-entropy loss between a given class and a uniform distribution covering its support. EOPE loss alleviates computational expenses associated with the estimation of the normalization term and RMSProp-like sampling procedure further accelerates training by reducing computational cost of sampling.

\section{Experiments}
\label{sec:test}
We evaluate proposed methods on the following data sets: MNIST \citep{mnist}, CIFAR~\citep{cifar}, KDD-99 \citep{KDD99}, Omniglot \citep{omniglot}, SUSY and HIGGS \citep{higgs}.
In order to reflect assumptions behind our approach we derive multiple tasks from each data set by varying size of the anomalous subset.

As the proposed methods target problems intermediate between one-class and two-class problems, we compare our approaches against the following algorithms:
\begin{itemize}
    \item conventional two-class classification with the cross-entropy loss;
    \item a semi-supervised method: dimensionality reduction by a deep AutoEncoder followed by a classifier with the cross-entropy loss;
    \item one-class methods: Deep SVDD \citep{doc} and Robust AutoEncoder \citep{rae}.
\end{itemize}

Since not all of the evaluated algorithms allow for a probabilistic interpretation, ROC AUC metric is reported.
As performance of certain algorithms (especially, two-class classification) varies significantly depending on the choice of negative class, we run each experiment multiple times, and report average and standard deviation of the metrics. The results are reported in Tables \ref{tab:higgs}, \ref{tab:susy}, \ref{tab:kdd}, \ref{tab:mnist}, \ref{tab:cifar}, \ref{tab:omniglot}. Detailed description of the experimental setup can be found in Appendix A.

In these tables, columns represent tasks with varying numbers of negative samples presented in the training set: numbers in the header indicate either number of classes that form a negative class (in case of MNIST, CIFAR, Omniglot and KDD data sets), or number of negative samples used (HIGGS and SUSY); `one-class' denotes absence of known anomalious samples. As one-class algorithms do not take into account negative samples, results of these are repeated for the tasks with known anomalies.

\begin{figure}[ptb]
\centering
\footnotesize
\begin{tabular}{|l|l|l|l|l|l|}
\hline & one class & 100 & 1000 & 10000 & 1000000 \\ \hline
 Robust AE & $0.530 \pm 0.002$ & $0.530 \pm 0.002$ & $0.530 \pm 0.002$ & $0.530 \pm 0.002$ & $0.530 \pm 0.002$ \\ \hline
 Deep SVDD & $0.497 \pm 0.006$ & $0.497 \pm 0.006$ & $0.497 \pm 0.006$ & $0.497 \pm 0.006$ & $0.497 \pm 0.006$ \\ \hline
 cross-entropy & - & $0.496 \pm 0.017$ & $0.529 \pm 0.007$ & $0.566 \pm 0.006$ & $0.858 \pm 0.002$ \\ \hline
 semi-supervised & - & $0.498 \pm 0.003$ & $0.522 \pm 0.003$ & $0.603 \pm 0.002$ & $0.745 \pm 0.005$ \\ \hline
 brute-force OPE & $0.499 \pm 0.009$ & $0.500 \pm 0.009$ & $0.520 \pm 0.003$ & $0.572 \pm 0.005$ & $0.859 \pm 0.001$ \\ \hline
 HMC EOPE & $0.491 \pm 0.000$ & $0.523 \pm 0.005$ & $\mathbf{0.567} \pm 0.008$ & $\mathbf{0.648} \pm 0.005$ & $0.848 \pm 0.001$ \\ \hline
 RMSProp EOPE & $0.498 \pm 0.002$ & $0.494 \pm 0.008$ & $0.531 \pm 0.008$ & $0.593 \pm 0.011$ & $\mathbf{0.861} \pm 0.000$ \\ \hline
 Deep EOPE & $\mathbf{0.531} \pm 0.000$ & $\mathbf{0.537} \pm 0.011$ & $0.560 \pm 0.008$ & $0.628 \pm 0.005$ & $0.860 \pm 0.001$ \\ \hline
\end{tabular}
\caption{Results on HIGGS data set. The first row indicates numbers of negative samples used in training.}
\label{tab:higgs}
\end{figure}

\begin{figure}[ptb]
\centering
\footnotesize
\begin{tabular}{|l|l|l|l|l|l|}
\hline & one class & 100 & 1000 & 10000 & 1000000 \\ \hline
 Robust AE & $0.394 \pm 0.012$ & $0.394 \pm 0.012$ & $0.394 \pm 0.012$ & $0.394 \pm 0.012$ & $0.394 \pm 0.012$ \\ \hline
 Deep SVDD & $0.541 \pm 0.022$ & $0.541 \pm 0.022$ & $0.541 \pm 0.022$ & $0.541 \pm 0.022$ & $0.541 \pm 0.022$ \\ \hline
 cross-entropy & - & $0.658 \pm 0.033$ & $0.736 \pm 0.021$ & $0.757 \pm 0.036$ & $0.871 \pm 0.006$ \\ \hline
 semi-supervised & - & $0.715 \pm 0.020$ & $0.766 \pm 0.009$ & $\mathbf{0.847} \pm 0.002$ & $0.876 \pm 0.000$ \\ \hline
 brute-force OPE & $\mathbf{0.648} \pm 0.035$ & $0.678 \pm 0.025$ & $0.729 \pm 0.029$ & $0.757 \pm 0.036$ & $0.871 \pm 0.006$ \\ \hline
 HMC EOPE & $0.472 \pm 0.000$ & $\mathbf{0.738} \pm 0.019$ & $\mathbf{0.770} \pm 0.012$ & $0.816 \pm 0.006$ & $0.877 \pm 0.000$ \\ \hline
 RMSProp EOPE & $0.443 \pm 0.038$ & $0.714 \pm 0.019$ & $0.760 \pm 0.016$ & $0.807 \pm 0.004$ & $0.877 \pm 0.000$ \\ \hline
 Deep EOPE & $0.468 \pm 0.118$ & $0.670 \pm 0.054$ & $0.746 \pm 0.024$ & $0.813 \pm 0.003$ & $\mathbf{0.878} \pm 0.000$ \\ \hline
\end{tabular}
\caption{Results on SUSY data set. The first row indicates numbers of negative samples used in training.}
\label{tab:susy}
\end{figure}

\begin{figure}[ptb]
\centering
\footnotesize
\begin{tabular}{|l|l|l|l|l|l|}
\hline & one class & 1 & 2 & 4 & 8 \\ \hline
 Robust AE & $\mathbf{0.972} \pm 0.006$ & $\mathbf{0.972} \pm 0.006$ & $\mathbf{0.972} \pm 0.006$ & $\mathbf{0.972} \pm 0.006$ & $\mathbf{0.972} \pm 0.006$ \\ \hline
 Deep SVDD & $0.939 \pm 0.014$ & $0.939 \pm 0.014$ & $0.939 \pm 0.014$ & $0.939 \pm 0.014$ & $0.939 \pm 0.014$ \\ \hline
 cross-entropy & - & $0.571 \pm 0.213$ & $0.300 \pm 0.182$ & $0.687 \pm 0.268$ & $0.619 \pm 0.257$ \\ \hline
 semi-supervised & - & $0.315 \pm 0.258$ & $0.469 \pm 0.286$ & $0.758 \pm 0.171$ & $0.865 \pm 0.087$ \\ \hline
 brute-force OPE & $0.398 \pm 0.108$ & $0.667 \pm 0.175$ & $0.394 \pm 0.261$ & $0.737 \pm 0.187$ & $0.541 \pm 0.257$ \\ \hline
 HMC EOPE & $0.786 \pm 0.200$ & $0.885 \pm 0.152$ & $0.919 \pm 0.055$ & $0.863 \pm 0.094$ & $0.958 \pm 0.023$ \\ \hline
 RMSProp EOPE & $0.765 \pm 0.216$ & $0.824 \pm 0.237$ & $0.770 \pm 0.213$ & $0.941 \pm 0.048$ & $0.960 \pm 0.021$ \\ \hline
 Deep EOPE & $0.602 \pm 0.279$ & $0.767 \pm 0.245$ & $0.548 \pm 0.279$ & $0.763 \pm 0.217$ & $0.786 \pm 0.267$ \\ \hline
\end{tabular}
\caption{Results on KDD-99 data set. The first row indicates numbers of original classes selected as negative class, at most 1000 examples are sampled from each original class.}
\label{tab:kdd}
\end{figure}

\begin{figure}[ptb]
\centering
\footnotesize
\begin{tabular}{|l|l|l|l|l|}
\hline & one class & 1 & 2 & 4 \\ \hline
 Robust AE & $\mathbf{0.978} \pm 0.017$ & $\mathbf{0.978} \pm 0.017$ & $0.978 \pm 0.017$ & $0.978 \pm 0.017$ \\ \hline
 Deep SVDD & $0.641 \pm 0.086$ & $0.641 \pm 0.086$ & $0.641 \pm 0.086$ & $0.641 \pm 0.086$ \\ \hline
 cross-entropy & - & $0.879 \pm 0.108$ & $0.957 \pm 0.050$ & $0.987 \pm 0.014$ \\ \hline
 semi-supervised & - & $0.934 \pm 0.035$ & $0.964 \pm 0.032$ & $0.984 \pm 0.012$ \\ \hline
 brute-force OPE & $0.786 \pm 0.112$ & $0.915 \pm 0.096$ & $0.968 \pm 0.041$ & $0.986 \pm 0.015$ \\ \hline
 HMC EOPE & $0.694 \pm 0.167$ & $0.933 \pm 0.060$ & $0.974 \pm 0.023$ & $0.989 \pm 0.011$ \\ \hline
 RMSProp EOPE & $0.720 \pm 0.186$ & $0.933 \pm 0.062$ & $0.977 \pm 0.023$ & $0.990 \pm 0.009$ \\ \hline
 Deep EOPE & $0.793 \pm 0.129$ & $0.942 \pm 0.048$ & $\mathbf{0.979} \pm 0.016$ & $\mathbf{0.991} \pm 0.007$ \\ \hline
\end{tabular}
\caption{Results on MNIST data set. The first row indicates numbers of original classes selected as negative class, 10 images are sampled from each original class.}
\label{tab:mnist}
\end{figure}

\begin{figure}[ptb]
\centering
\footnotesize
\begin{tabular}{|l|l|l|l|l|}
\hline & one class & 1 & 2 & 4 \\ \hline
 Robust AE & $\mathbf{0.585} \pm 0.126$ & $0.585 \pm 0.126$ & $0.585 \pm 0.126$ & $0.585 \pm 0.126$ \\ \hline
 Deep SVDD & $0.546 \pm 0.058$ & $0.546 \pm 0.058$ & $0.546 \pm 0.058$ & $0.546 \pm 0.058$ \\ \hline
 cross-entropy & - & $0.659 \pm 0.093$ & $0.708 \pm 0.086$ & $0.748 \pm 0.082$ \\ \hline
 semi-supervised & - & $0.587 \pm 0.109$ & $0.634 \pm 0.109$ & $0.671 \pm 0.093$ \\ \hline
 brute-force OPE & $0.549 \pm 0.098$ & $\mathbf{0.688} \pm 0.087$ & $\mathbf{0.719} \pm 0.079$ & $\mathbf{0.757} \pm 0.073$ \\ \hline
 HMC EOPE & $0.547 \pm 0.116$ & $0.678 \pm 0.091$ & $0.709 \pm 0.084$ & $0.739 \pm 0.074$ \\ \hline
 RMSProp EOPE & $0.565 \pm 0.111$ & $0.678 \pm 0.081$ & $0.715 \pm 0.083$ & $0.746 \pm 0.069$ \\ \hline
 Deep EOPE & $0.564 \pm 0.094$ & $0.674 \pm 0.100$ & $0.690 \pm 0.092$ & $0.719 \pm 0.099$ \\ \hline
\end{tabular}
\caption{Results on CIFAR-10 data set. The first row indicates numbers of original classes selected as negative class, 10 images are sampled from each original class.}
\label{tab:cifar}
\end{figure}

\begin{figure}[ptb]
\centering
\footnotesize
\begin{tabular}{|l|l|l|l|l|}
\hline & one class & 1 & 2 & 4 \\ \hline
 Robust AE & $\mathbf{0.771} \pm 0.221$ & $0.771 \pm 0.221$ & $0.771 \pm 0.221$ & $0.771 \pm 0.221$ \\ \hline
 Deep SVDD & $0.640 \pm 0.153$ & $0.640 \pm 0.153$ & $0.640 \pm 0.153$ & $0.640 \pm 0.153$ \\ \hline
 cross-entropy & - & $0.799 \pm 0.162$ & $\mathbf{0.862} \pm 0.115$ & $0.855 \pm 0.125$ \\ \hline
 semi-supervised & - & $0.737 \pm 0.134$ & $0.821 \pm 0.104$ & $0.805 \pm 0.121$ \\ \hline
 brute-force OPE & $0.591 \pm 0.161$ & $0.724 \pm 0.222$ & $0.765 \pm 0.208$ & $0.825 \pm 0.126$ \\ \hline
 HMC EOPE & $0.710 \pm 0.178$ & $0.801 \pm 0.139$ & $0.842 \pm 0.112$ & $0.842 \pm 0.115$ \\ \hline
 RMSProp EOPE & $0.678 \pm 0.274$ & $\mathbf{0.821} \pm 0.143$ & $0.855 \pm 0.112$ & $\mathbf{0.863} \pm 0.111$ \\ \hline
 Deep EOPE & $0.696 \pm 0.172$ & $0.808 \pm 0.140$ & $0.851 \pm 0.110$ & $0.842 \pm 0.122$ \\ \hline
\end{tabular}
\caption{Results on Omniglot data set. The first row indicates numbers of original classes selected as negative class, 10 images are sampled from each original class. Greek, Braille and Futurama alphabets are used as normal classes.}
\label{tab:omniglot}
\end{figure}

In our experiments, we make several observations. Firstly, proposed methods generally outperform baseline methods, especially on the problems with a significant overlap between classes (SUSY, HIGGS and, possibly, CIFAR), and consistently show comparable performance on test problems.
Secondly, we observe increasing performance as more negative samples are included in training set, while being consistently above or similar to that of conventional two-class classification. Lastly, to our surprise, brute-force OPE performs relatively well even on high-dimensional problems, which might indicate that gradients produced by its regularization term have variance sufficiently low for a proper convergence.

The main drawback of the OPE and EOPE methods is a slow training, which is largely due to usage of Monte-Carlo methods. It is partially alleviated by fast approximation of Hamiltonian Monte-Carlo and usage of a generator \citep{kim2016deep}, and can potentially be improved further, by advanced Monte-Carlo techniques \citep[for example, ][]{levy2017generalizing}.

\section{Conclusion}
\label{sec:conclusion}

We present a new family of anomaly detection algorithms which can be efficiently applied to the problems intermediate between one-class and two-class settings. Solutions produced by these methods combine the best features of one-class and two-class approaches. In contrast to conventional one-class approaches, proposed methods can effectively utilise any number of known anomalous examples, and, unlike conventional two-class classification, does not require a representative sample of anomalous data.
Our experiments show better or comparable performance to conventional two-class and one-class algorithms. Our approach is especially beneficial for anomaly detection problems, in which anomalous data is non-representative, or might evolve over time.

\acks{The research leading to these results has received funding from Russian Science Foundation under grant agreement n$^{\circ}$ 17-72-20127.}

\bibliography{main.bib}

\begin{thebibliography}{23}
\providecommand{\natexlab}[1]{#1}
\providecommand{\url}[1]{\texttt{#1}}
\expandafter\ifx\csname urlstyle\endcsname\relax
  \providecommand{\doi}[1]{doi: #1}\else
  \providecommand{\doi}{doi: \begingroup \urlstyle{rm}\Url}\fi

\bibitem[KDD(1999)]{KDD99}
\textrm{KDD} cup 1999 dataset: Intrusion detection system, 1999.
\newblock URL \url{https://archive.ics.uci.edu/ml/datasets/kdd+cup+1999+data}.

\bibitem[Baldi et~al.(2014)Baldi, Sadowski, and Whiteson]{higgs}
Pierre Baldi, Peter Sadowski, and Daniel Whiteson.
\newblock Searching for exotic particles in high-energy physics with deep
  learning.
\newblock \emph{Nature communications}, 5:\penalty0 4308, 2014.

\bibitem[Bengio et~al.(2009)]{bengio2009learning}
Yoshua Bengio et~al.
\newblock Learning deep architectures for ai.
\newblock \emph{Foundations and trends{\textregistered} in Machine Learning},
  2\penalty0 (1):\penalty0 1--127, 2009.

\bibitem[Chalapathy et~al.(2018)Chalapathy, Menon, and Chawla]{ocnn}
Raghavendra Chalapathy, Aditya~Krishna Menon, and Sanjay Chawla.
\newblock Anomaly detection using one-class neural networks.
\newblock \emph{arXiv preprint arXiv:1802.06360}, 2018.

\bibitem[Duane et~al.(1987)Duane, Kennedy, Pendleton, and Roweth]{hmc}
Simon Duane, Anthony~D Kennedy, Brian~J Pendleton, and Duncan Roweth.
\newblock Hybrid monte carlo.
\newblock \emph{Physics letters B}, 195\penalty0 (2):\penalty0 216--222, 1987.

\bibitem[Jin et~al.(2017)Jin, Lazarow, and Tu]{icn}
Long Jin, Justin Lazarow, and Zhuowen Tu.
\newblock Introspective classification with convolutional nets.
\newblock In \emph{Advances in Neural Information Processing Systems}, pages
  823--833, 2017.

\bibitem[Kim and Bengio(2016)]{kim2016deep}
Taesup Kim and Yoshua Bengio.
\newblock Deep directed generative models with energy-based probability
  estimation, 2016.

\bibitem[Kingma and Ba(2014)]{kingma2014adam}
Diederik~P Kingma and Jimmy Ba.
\newblock Adam: A method for stochastic optimization.
\newblock \emph{arXiv preprint arXiv:1412.6980}, 2014.

\bibitem[Krizhevsky and Hinton(2009)]{cifar}
Alex Krizhevsky and Geoffrey Hinton.
\newblock Learning multiple layers of features from tiny images.
\newblock Technical report, Citeseer, 2009.

\bibitem[Lake et~al.(2015)Lake, Salakhutdinov, and Tenenbaum]{omniglot}
Brenden~M Lake, Ruslan Salakhutdinov, and Joshua~B Tenenbaum.
\newblock Human-level concept learning through probabilistic program induction.
\newblock \emph{Science}, 350\penalty0 (6266):\penalty0 1332--1338, 2015.

\bibitem[LeCun et~al.(1998)LeCun, Bottou, Bengio, Haffner, et~al.]{mnist}
Yann LeCun, L{\'e}on Bottou, Yoshua Bengio, Patrick Haffner, et~al.
\newblock Gradient-based learning applied to document recognition.
\newblock \emph{Proceedings of the IEEE}, 86\penalty0 (11):\penalty0
  2278--2324, 1998.

\bibitem[Levy et~al.(2017)Levy, Hoffman, and
  Sohl-Dickstein]{levy2017generalizing}
Daniel Levy, Matthew~D Hoffman, and Jascha Sohl-Dickstein.
\newblock Generalizing hamiltonian monte carlo with neural networks.
\newblock \emph{arXiv preprint arXiv:1711.09268}, 2017.

\bibitem[Liu et~al.(2008)Liu, Ting, and Zhou]{liu2008isolation}
Fei~Tony Liu, Kai~Ming Ting, and Zhi-Hua Zhou.
\newblock Isolation forest.
\newblock In \emph{2008 Eighth IEEE International Conference on Data Mining},
  pages 413--422. IEEE, 2008.

\bibitem[Ruff et~al.(2018)Ruff, G{\"o}rnitz, Deecke, Siddiqui, Vandermeulen,
  Binder, M{\"u}ller, and Kloft]{doc}
Lukas Ruff, Nico G{\"o}rnitz, Lucas Deecke, Shoaib~Ahmed Siddiqui, Robert
  Vandermeulen, Alexander Binder, Emmanuel M{\"u}ller, and Marius Kloft.
\newblock Deep one-class classification.
\newblock In \emph{International Conference on Machine Learning}, pages
  4390--4399, 2018.

\bibitem[Sch{\"o}lkopf et~al.(2000)Sch{\"o}lkopf, Williamson, Smola,
  Shawe-Taylor, and Platt]{scholkopf2000support}
Bernhard Sch{\"o}lkopf, Robert~C Williamson, Alex~J Smola, John Shawe-Taylor,
  and John~C Platt.
\newblock Support vector method for novelty detection.
\newblock In \emph{Advances in neural information processing systems}, pages
  582--588, 2000.

\bibitem[Schölkopf and Smola(2002)]{SVM}
B.~Schölkopf and A.J. Smola.
\newblock \emph{Support vector machines, regularization, optimization, and
  beyond}.
\newblock MIT Press, 2002.

\bibitem[Simonyan and Zisserman(2014)]{vgg}
Karen Simonyan and Andrew Zisserman.
\newblock Very deep convolutional networks for large-scale image recognition.
\newblock \emph{arXiv preprint arXiv:1409.1556}, 2014.

\bibitem[Tax and Duin(2001)]{svdd}
David~MJ Tax and Robert~PW Duin.
\newblock Uniform object generation for optimizing one-class classifiers.
\newblock \emph{Journal of machine learning research}, 2\penalty0
  (Dec):\penalty0 155--173, 2001.

\bibitem[Tieleman(2008)]{tieleman2008training}
Tijmen Tieleman.
\newblock Training restricted boltzmann machines using approximations to the
  likelihood gradient.
\newblock In \emph{Proceedings of the 25th international conference on Machine
  learning}, pages 1064--1071. ACM, 2008.

\bibitem[Tieleman and Hinton(2009)]{prbm}
Tijmen Tieleman and Geoffrey Hinton.
\newblock Using fast weights to improve persistent contrastive divergence.
\newblock In \emph{Proceedings of the 26th Annual International Conference on
  Machine Learning}, pages 1033--1040. ACM, 2009.

\bibitem[Tieleman and Hinton(2012)]{rmsprop}
Tijmen Tieleman and Geoffrey Hinton.
\newblock Lecture 6.5-rmsprop: Divide the gradient by a running average of its
  recent magnitude.
\newblock \emph{COURSERA: Neural networks for machine learning}, 4\penalty0
  (2):\penalty0 26--31, 2012.

\bibitem[Tu(2007)]{dgl}
Zhuowen Tu.
\newblock Learning generative models via discriminative approaches.
\newblock In \emph{2007 IEEE Conference on Computer Vision and Pattern
  Recognition}, pages 1--8. IEEE, 2007.

\bibitem[Zhou and Paffenroth(2017)]{rae}
Chong Zhou and Randy~C Paffenroth.
\newblock Anomaly detection with robust deep autoencoders.
\newblock In \emph{Proceedings of the 23rd ACM SIGKDD International Conference
  on Knowledge Discovery and Data Mining}, pages 665--674. ACM, 2017.

\end{thebibliography}

\newpage
\appendix

\section*{Appendix A.}
\label{sec:results}
This section provides detailed description of the experimental setup.

In order to make a clear comparison between methods, network architectures are made as close as possible. For image data (MNIST, CIFAR, Omniglot) VGG-like networks (\citep{vgg}) are used, for tabular data 5-layers dense networks are used\footnote{Implementation can be found at \url{https://gitlab.com/mborisyak/ope}.}.

We evaluate the following proposed methods:
\begin{itemize}
    \item brute-force OPE: described in Algorithm \ref{algo:ope};
    \item HMC EOPE: Algorithm \ref{algo:eope} equipped with Hamiltonian Monte-Carlo;
    \item RMProp EOPE: Algorithm \ref{algo:eope} with the pseudo-MCMC described by Equations~\ref{eq:rmsprop1}~and~\ref{eq:rmsprop2}.
    \item Deep EOPE: Algorithm \ref{algo:eope} with MCMC sampling procedure replaced by a generator as in \citep{kim2016deep}.
\end{itemize}

All OPE and EOPE models are trained with $\varepsilon = 0.95$.
All MCMC chains are persistent (by analogy with \citep{prbm}) and 4 MCMC steps are performed for each gradient step.
All networks are optimized by adam algorithm (\citep{kingma2014adam}) with learning rate $5 \cdot 10^{-4}$, $\beta_1=0.9$, $\beta_2=0.999$.

In order to reflect assumptions behind proposed methods, we derive several tasks from each original data set considered.
For SUSY, HIGGS and KDD-99 data set positive class is fixed according to data sets' descriptions; for MNIST and CIFAR-10 data sets each class is considered as positive; for Omniglot data set we choose `Braille', `Futurama' and `Greek' alphabets are chosen as positive classes.

In order to fully demonstrate advantages of OPE and EOPE methods we vary sample sizes for negative class: for SUSY and HIGGS data sets only a small number of negative examples is randomly selected ($0$, $10^2$, $10^3$, $10^4$ and $10^5$); for multi-class data sets several classes are randomly selected (without replacement) and subsampled, for MNIST, CIFAR and Omniglot data sets $0$, $1$, $2$ and $4$ classes are selected with $10$ examples from each, for KDD-99 maximum number of samples per class is limited by $10^3$.

Original train-test splits are respected when possible (for SUSY and HIGGS data sets splits are random and fixed for all derived tasks) --- test sets are not modified in any way.

\section*{Appendix B.}
\label{sec:proof}
Here we provide a formal proof of Theorem 1 from the Section~\ref{sec:eope}.
For the sake of simplicity we split the proof into two lemmas.

\noindent
{\bf Lemma 1}{\it
    Let $(\mathcal{X}, \|\cdot\|)$ be a Banach space, $P(x)$ --- a continuous probability density function such that $\Omega = \supp P$ is an open set in $\mathcal{X}$.
    If continuous function $g^* : \Omega \to \mathbb{R}$ minimizes $\mathcal{L}^E_1$ (defined by Equation~\ref{eq:eocloss}) with $P(x \mid \mathcal{C}^+) = P(x)$, then there exists a strictly increasing function $s: \mathbb{R} \to \mathbb{R}$, such that $g^*(x) = s(P(x))$.} \hfill\BlackBox

\noindent
{\bf Proof}.
   Consider a continuous function $g: \Omega \to \mathbb{R}$.
    We show that if $g$ can not be represented as $s(P(x))$, then $g$ does not minimize $\mathcal{L}^E_1$. This is demonstrated by constructing another continuous function $g'$ that achieves lower loss than $g$.
    
    If $g$ can not be represented as $s(P(x))$ then a pair of points $x_1$ and $x_2$ can be found such that $P(x_1) < P(x_2)$ and $g(x_1) \geq g(x_2)$.

    Due to continuity of $P$ and $g$, it is possible to find such neighborhoods of $x_1$ and $x_2$, that the difference in probability densities remains large, while differences in values of $g$ become insignificant or negative. More formally, for every $\delta > 0$ there exists $r > 0$ such that open balls $B_1 = B(x_1, r)$ and $B_2 = B(x_2, r)$, $B_1, B_2 \subset \Omega$ satisfy following properties:
    \begin{eqnarray}
        \inf_{B_2} P - \sup_{B_1} P &>& \Delta; \label{eq:monviolation}\\
        \sup_{B_2} g - \inf_{B_1} g &<& \delta; \label{eq:gsoftgap}
    \end{eqnarray}
    where $2 \Delta = P(x_2) - P(x_1)$.

    We define function $g'_{\alpha, \beta}$ as $g'_{\alpha, \beta}(x) = g(x) - \alpha\,h(x - x_1) + \beta\,h(x - x_2)$, where $h: \mathcal{X} \to \mathbb{R}$, $\alpha, \beta > 0$; the exact form of $h$ is not important, nevertheless, for clarity, let
    \begin{equation}
        h(x) = r^{-1} \cdot \max(r -\|x\|, 0).
    \end{equation}
    We restrict our attention to such values of $\alpha$ and $\beta$, that $g'_{\alpha, \beta}$ has the same normalization constant as $g$:
    \begin{equation}
        \Delta Z(\alpha, \beta) = \int_\Omega \exp(g'_{\alpha, \beta}(x)) dx - \int_\Omega \exp(g(x)) dx = 0\label{eq:normpreservation}.
    \end{equation}
    
    Equation~\ref{eq:monviolation} implies that $B_1$ and $B_2$ do not intersect and, since $g(x) = g'_{\alpha, \beta}(x)$ for $x \in \Omega \setminus (B_1 \cup B_2)$, $\Delta Z(\alpha, \beta)$ consists of two non-zero terms:
    \begin{equation}
        \Delta Z(\alpha, \beta) = \Delta Z_1(\alpha) + \Delta Z_2(\beta); \nonumber
    \end{equation}
    where:
    \begin{eqnarray}
        \Delta Z_1(\alpha) &=& \int_{B_1} \exp(g(x) - \alpha h(x - x_1)) - \exp(g(x)) d x;\nonumber \\
        \Delta Z_2(\beta) &=& \int_{B_2} \exp(g(x) + \beta h(x - x_2)) - \exp(g(x)) d x. \nonumber
    \end{eqnarray}
    For every $\alpha \geq 0$ there exist a unique $\beta^*(\alpha) \geq 0$ such that $(\alpha, \beta^*(\alpha))$ is a solution for Equation~\ref{eq:normpreservation}. Notice also, that $\beta^*(\alpha)$ is a continuous, strictly increasing function and $\beta^*(0) = 0$.
    
    Notice, that for small values of $\alpha$ and $\beta$
    \begin{eqnarray}
        \Delta Z_1(\alpha) &=& \int_{B_1} -\alpha h(x - x_1) \exp(g(x)) + \mathcal{O}(\alpha^2 h^2(x - x_1)) d x;\nonumber \\
        \Delta Z_2(\beta) &=& \int_{B_2} \beta h(x - x_2) \exp(g(x)) + \mathcal{O}(\beta^2 h^2(x - x_2)) d x. \nonumber
    \end{eqnarray}
    therefore,
    \begin{equation}
        \lim_{\delta \to 0} \lim_{\alpha \to 0} \frac{\alpha}{\beta^*(\alpha)} \leq 1. \label{eq:abprop}
    \end{equation}

    Similarly to $\Delta Z(\alpha, \beta)$, $\Delta L^+(\alpha, \beta) = L^+(g'_{\alpha, \beta}) - L^+(g)$ can be split into two parts:
    \begin{eqnarray}
        \Delta L^+(\alpha, \beta) &=& \Delta L^+_1(\alpha) + \Delta L^+_2(\beta) \label{eq:lplus};\\
        \Delta L^+_1(\alpha) &=& \int_{B_1} P(x)\left[ l(g(x) - \alpha h(x - x_1)) - l(g(x))  \right] dx;\nonumber \\
        \Delta L^+_2(\beta) &=& \int_{B_2} P(x)\left[ l(g(x) + \beta h(x - x_2)) - l(g(x))  \right] dx; \nonumber
    \end{eqnarray}
    where $l(y) = \log(1 + \exp(-y))$.
    
    Note, that for a positive $\Delta y$
    \begin{equation}
        \frac{\Delta y}{1 + \exp(y + \Delta y)} < l(y) - l(y + \Delta y) < \frac{\Delta y}{1 + \exp(y)}; \nonumber
    \end{equation}
    therefore,
    \begin{eqnarray}
        \Delta L^+_1(\alpha) &<& \int_{\|\chi\| < r} \alpha h(\chi) P(x_1 + \chi) J_1(\alpha, \chi) d\chi; \nonumber\\
        \Delta L^+_2(\beta) &<& -\int_{\|\chi\| < r} \beta h(\chi) P(x_2 + \chi) J_2(\beta, \chi) d\chi. \nonumber
    \end{eqnarray}
    where:
    \begin{eqnarray}
        J_1(\alpha, \chi) &=& \frac{1}{1 + \exp(g(x_1 + \chi) - \alpha h(\chi))} \label{eq:j1};\\
        J_2(\beta, \chi) &=& \frac{1}{1 + \exp(g(x_2 + \chi) + \beta h(\chi))}\label{eq:j2};
    \end{eqnarray}
    
    hence,
    \begin{equation}
        \Delta L^+(\alpha, \beta) < \int_{\|\chi\| < r} h(\chi)\left[P(x_1 + \chi) \alpha J_1(\alpha, \chi) - P(x_2 + \chi)\beta J_2(\alpha, \chi)) \right] d\chi. \nonumber
    \end{equation}
    
    Note, that
    \begin{multline}
        P(x_1 + \chi)\alpha J_1(\alpha, \chi) - P(x_2 + \chi)\beta J_2(\beta, \chi) \leq \\
        \frac{\alpha P_1}{1 + \exp(G_1)} -  \frac{\beta P_2}{1 + \exp(G_2 + \beta)} \equiv J(\alpha, \beta); \nonumber
    \end{multline}
    where: $G_1 = \inf_{B_1} g$, $G_2 = \sup_{B_2} g$, $P_1 = \sup_{B_1} P$, $P_2 = \inf_{B_2} P$.
    
    Now, our aim is to prove that $J(\alpha, \beta) \leq 0$ has a solution in form $(\alpha, \beta^*(\alpha))$:
    \begin{equation}
        J(\alpha, \beta^*(\alpha)) < 0 \Leftrightarrow \frac{\alpha}{\beta^*(\alpha)} < C(\beta^*(\alpha)); \label{eq:betaineq1}
    \end{equation}
    where:
    \begin{equation}
        C(\beta) = \frac{P_2}{P_1} \frac{1 + \exp(G_1)}{1 + \exp(G_2 + \beta)}. \nonumber
    \end{equation}
    Note, that for each $0 < \delta < \log(P_2) - \log(P_1)$, and for each $0 < \beta < \log(P_2) - \log(P_1) - \delta$, $C(\beta) > 1$. In combination with Equation~\ref{eq:abprop}, this implies that Inequality~\ref{eq:betaineq1} is satisfied for some $\alpha > 0$ and $\beta = \beta^*(\alpha)$, therefore, $\Delta Z(\alpha, \beta) = 0$ and $\Delta L^+(\alpha, \beta) < 0$ are simultaneously satisfied for some $\alpha > 0$ and $\beta > 0$. This implies, that function $g'_{\alpha_0, \beta^*(\alpha_0)}$ has the same normalization constant $Z$ as the original one, and reduces value of $L^+$, hence, $g$ does not minimize $\mathcal{L}^E_1$, which concludes this proof. \hfill\BlackBox

\vskip 0.2in
\noindent
{\bf Lemma 2.} {\it
    For every function $s$ that satisfies Lemma 1:
    \begin{equation}
        \inf_\Omega P = 0 \Rightarrow \lim_{y \to 0} s(y) = -\infty. \nonumber
    \end{equation} } \hfill\BlackBox

\noindent
{\bf Proof}.
    Suppose that $\lim_{y \to 0} s(y) = S \in \mathbb{R}$.
    
    For every sufficiently small $\Delta > 0$, we can pick points $x_1, x_2 \in \Omega$, radius $r > 0$ and two open balls $B_1(x_1, r)$, $B_2(x_2, r)$ such that
    \begin{eqnarray}
        \sup_{B_1} P &<& \Delta ;\nonumber \\
        \sup_{B_2} P &>& 8 \Delta \nonumber.
    \end{eqnarray}
    
    Now we can introduce the same definitions and constructs as in Lemma 1, applied for $B_1$, $B_2$ and $g$.
    Consider $\alpha > 0$, such that $\alpha < 2 \beta^*(\alpha)$ (such values always exist due to Equation~\ref{eq:abprop}).
    Note that since $\inf_\Omega g \geq S$, for every $\beta > 0$, $J_2$ (defined by Equation~\ref{eq:j2}) is bounded from below
    \begin{equation}
        \inf_{\|\chi\| < r} J_2(\beta, \chi) \geq \frac{1}{1 + \exp(S + \beta)}. \nonumber
    \end{equation}
    
    Consider
    \begin{equation}
        \Delta = \min\left[ \frac{\sup_{\|\chi\| < r} J_1(\alpha, \chi)}{\inf_{\|\chi\| < r} J_2(\beta^*(\alpha), \chi)}, 1 \right] \cdot \frac{\sup_{B_2} P}{4}. \nonumber
    \end{equation}
    Such choice of $\Delta$ guarantees that for every $\chi$, such that $\|\chi\| < r$,
    \begin{equation}
        \alpha P(x_1  + \chi) J_1(\alpha, \chi) < \beta^*(\alpha) P(x_2 + \chi) J_2(\beta^*(\alpha), \chi); \nonumber
    \end{equation}
    where $J_1$ and $J_2$ are defined by Equations~\ref{eq:j1}~and~\ref{eq:j2}. This makes $\Delta L^+$ from Equation~\ref{eq:lplus} negative, which, in turn, implies that $g$ does not minimize $\mathcal{L}^E_1$, which contradicts our assumptions. \hfill\BlackBox
\end{document}